\documentclass[letterpaper]{article}
\usepackage{iccc}

\usepackage{times}
\usepackage{helvet}
\usepackage{courier}
\usepackage{csquotes} 
\usepackage{xcolor}
\usepackage{natbib} 
\usepackage{graphicx}
\usepackage{rotating}
\usepackage{booktabs} 
\usepackage{multirow} 
\newcommand*\rot{\rotatebox{90}} 
\usepackage{wasysym}
\usepackage{flushend}
\usepackage{acronym}
\usepackage{soul}
\usepackage{fixltx2e}
\usepackage{booktabs}
\newacro{CC}{computational creativity}
\newacro{ICCC}{International Conference on Computational Creativity}

\title{Embodiment and Computational Creativity}
\author{Christian Guckelsberger\textsuperscript{1--3,$\dagger$}, Anna Kantosalo\textsuperscript{2,$\dagger$}, Santiago Negrete-Yankelevich\textsuperscript{4,$\ddagger$} \and Tapio Takala\textsuperscript{2,$\ddagger$}\\
\textsuperscript{1}Finnish Center for Artificial Intelligence\\ 
\textsuperscript{2}Department of Computer Science, Aalto University, Espoo, Finland\\
\textsuperscript{3}School of Electronic Engineering and Computer Science, Queen Mary University of London, London, UK\\
\textsuperscript{4} Department of Information Technology, Metropolitan Autonomous University (Cuajimalpa), Mexico City, Mexico\\
\textsuperscript{$\dagger$}Contributed equally and share first authorship \textsuperscript{$\ddagger$}Contributed equally and share second authorship\\
christian.guckelsberger@aalto.fi, anna.kantosalo@aalto.fi, snegrete@cua.uam.mx, tta@cs.hut.fi\\
}

\begin{document} 
\maketitle

\begin{abstract}
\begin{quote}
We conjecture that creativity and the perception of creativity are, at least to some extent, shaped by embodiment. This makes embodiment highly relevant for \ac{CC} research, but existing research is scarce and the use of the concept highly ambiguous. We overcome this situation by means of a systematic review and a prescriptive analysis of publications at the International Conference on Computational Creativity. We adopt and extend an established typology of embodiment to resolve ambiguity through identifying and comparing different usages of the concept. We collect, contextualise and highlight opportunities and challenges in embracing embodiment in \ac{CC} as a reference for research, and put forward important directions to further the embodied \ac{CC} research programme.
\end{quote}
\end{abstract}

\section{Introduction}

Most researchers agree that \emph{creativity} and \emph{intelligence} are closely intertwined \citep{kaufmanCreativityIntelligence}. Moreover, it is widely accepted that intelligence is conditioned on \emph{embodiment} \citep{brooks1991intelligence, clark1998being}; with the words of \citeauthor{pfeifer2001understanding}, \enquote{intelligence cannot merely exist in the form of an  abstract algorithm but requires a physical instantiation, a body} \citep[][p.~649]{pfeifer2001understanding}. Without insisting on physicality, we conjecture that creativity, as a form of intelligent cognition, is also shaped by embodiment.

This makes embodiment highly relevant across the entire continuum of \acf{CC} research \citep{perez2018computational, veale2019systematizing}, from engineering artificial systems that can be considered autonomously creative \citep{colton2008creativity, Colton2012}, to understanding creativity in living beings through computational modelling and simulation \citep{boden2003creative}. Here, some of the most striking questions are if reproducing human-like creativity by computational means is at all possible without also reproducing human embodiment \citep{Guckelsberger2017a, valverdenegrete2017}, and how changes to a system's embodiment affect its potential creativity.

Unlike psychologists, whose human subjects share similar embodiments, \ac{CC} researchers can go beyond investigating the effect of embodiment on \emph{creativity per se} and explore how people's \emph{perception of creativity} \citep{colton2008creativity,colton2018issues} as exhibited by an artificial system is affected by their own and the system's embodiment. If such an effect existed, embodiment would have to be considered an integral factor in the evaluation of \ac{CC} \citep{jordanous2012standardised} to facilitate fairer comparisons between computational systems and with human creativity. Moreover, using this knowledge, researchers could tune a system's embodiment to improve the perception of its creativity, and, vice versa, how it perceives the creativity of others.

Given these potential ramifications, it is surprising and alarming that \enquote{embodiment} seems to have received little attention in \ac{CC} research. One potential reason for the apparent void is that the very concept is highly ambiguous \citep{ziemke2003s}. Theories of embodied cognition -- from minimal accounts that \enquote{rule out anatomy and bodily movement as important}~\citep{gallagher2011interpretations}, to radical approaches that understand cognition as inextricably bound to bodily processes -- conceptualise embodiment differently. Moreover, embodied cognition is closely associated with other, popular extra-cranial and extra-bodily theories of cognition, in particular theories of enactive, embedded and extended cognition. Due to their proximity and interdependence, they are frequently grouped together into the complex of \enquote{4E cognition}.

Crucially though, there appears to be very little awareness of this ambiguity and complexity within \ac{CC} research. At present, it is perfectly imaginable that two researchers excitedly referred to the \enquote{embodiment} of their respective system without noticing that they are talking about entirely different things. We consider this a major problem, given that the various types of embodiment likely have radically different effects on (the perception of) creativity. We argue that the advancement of the field through concerted investigations requires the use of common definitions of embodiment.

The first goal of this paper is to counteract this ambiguity and provide a rich overview of what some authors have already coined \enquote{embodied \acf{CC}} \citep{saunders2014accomplice, Guckelsberger2017a, colton2018issues} research. Based on a systematic review of related work at the \acf{ICCC} as the prime and domain-agnostic venue of \ac{CC} research, we answer the following research questions:
\begin{description}
\item[RQ1:]What types of embodiment have been embraced in \ac{CC} research, and how has the usage evolved over time?
\item[RQ2:]Why did \ac{CC} researchers embrace these embodiment types in their work, and what challenges did they face?
\item[RQ3:]What does \ac{CC} research reveal about the relationship of embodiment and (the perception of) creativity?
\end{description}
To counteract ambiguity in the usage of the embodiment concept, we extend and apply a well established typology of embodiment informed by cognitive science to assess the specific types addressed in each relevant contribution. By making transparent which types of embodiment have been embraced, and by highlighting our challenges in assessing them, we want to provide a frame of reference for researchers to adequately and unambiguously address questions of embodiment in their work. Our insights moreover allow us to to provide recommendations for an embodied \ac{CC} research programme -- the second goal of this paper.

\section{Types of Embodiment}

Much research focuses on distinguishing theories of embodied cognition \citep[e.g.][]{gallagher2011interpretations}, but comparisons of the underlying and varying conceptualisations of embodiment are rare. To disambiguate different uses of the embodiment concept in \ac{CC}, we adopt and extend the well-established typology by \citet{ziemke2003s}, who distinguishes six types of embodiment informed by research in cognitive science and robotics. We introduce three additions to this typology (one additional type, two additions to existing types) based on more recent insights, and highlight them in \emph{italics} below.
\begin{description}
\item{\textbf{structural coupling}}, characterising systems that can perturbate, and, vice versa, be perturbated by their surrounding environment \citep{varela1991embodied}. Such perturbations facilitate a minimal interaction between the system and environment, in which each has the potential to affect the other's state \citep{quick1999bots}.
\item{\textbf{historical}, characterising systems whose present state is the result of a history of structural couplings, developed through interactions with the environment over time  \citep{varela1991embodied, ziemke1999rethinking}.} 
\item{\textbf{virtual}}, \emph{characterising simulated systems embedded in and distinguished from a simulated environment. The virtual body can act on the environment and vice versa.}
\item{\textbf{physical}}, characterising systems with a physical body \citep{brooks1990elephants, pfeifer2001understanding} that can interact with the environment by being subjected to and by exercising physical force. Most prominently, robots are physically embodied \citep{pfeifer2005new}.
\item{\textbf{organismoid}}, characterising \emph{virtually} or physically embodied systems with the same or a similar shape and sensorimotor equipment as living organisms, e.g.~animals. We consider \textbf{humanoid} embodiment as approximations of the human body a subset of organismoid embodiment.
\item{\textbf{organismic}}, applying to living \emph{and artificial} systems capable of organisational closure, i.e.~of maintaining their organisation and surrounding boundary against internal and external perturbations by means of self-producing processes \citep{von1920theoretische, maturana1987tree}. A prominent, minimal example is the living cell which, in a self-referential process, maintains its organisation, including its membrane, against perturbations from the surrounding environment \citep{agmon2016structure}.
\end{description}

We briefly justify our additions. We have complemented \emph{physical} with \emph{virtual embodiment}, because AI researchers have successfully reproduced (super-)human cognitive abilities in virtual agents, embedded in e.g.~high fidelity physics simulations \citep{lillicrap2015continuous} or, often more coarse, videogame worlds \citep{mnih2015human}. Applying AI techniques to virtual agents in simulated worlds rather than to physically embodied systems allows for scalability, incremental development, and rapid iteration, amongst other advantages \citep{kiela2016virtual}. As a corollary, we have extended \emph{organismoid} embodiment to \emph{virtually embodied} systems, e.g.~in the form of believable game characters with a human or animal-like appearance.
We have finally extended \emph{organismic embodiment} to artificial systems. 
Originally restricted to the biochemical domain, this type required the capacity for autopoiesis, i.e. self-production, to facilitate a radical form of autonomy. \citeauthor{varela1979principles} overcomes this limitation by introducing the concept of organisational closure as \enquote{operational characterization of autonomy in general, living or otherwise}~\citep{varela1979principles}. \citet{froese2009enactive} advocate that organisational closure can be realised by AI systems, and survey existing examples. 
We can thus consider organismic embodiment in artificial systems, which makes it relevant for \ac{CC} research. Although organismic embodiment is arguably the least well-established type, we include it for its presence in existing \ac{CC} theory \citep{saunders2012towards,Guckelsberger2017a}, its potential future implications for \ac{CC}, and its central role in related debates, e.g.~on agency~\citep{polani2016towards}.

In our adaptation of Ziemke's \citeyearpar{ziemke2003s} typology, we have dropped what e.g.~\citet{dautenhahn1997could} and \citet{barsalou2003social} refer to as social embodiment, because it denotes the use of different types of embodiment to facilitate social interaction \citep{ziemke2003s}, and is thus orthogonal to, and not at the same \enquote{atomic} level, as the other types. \citet{metzinger2014first} has proposed a distinction between 1st, 2nd and 3rd order embodiment based on a system's computational abilities, corresponding to (1) physical, reactive systems without explicit computation, (2) systems that explicitly represent themselves as embodied agents, and (3) systems that can consciously experience some of these body representations. We disregard this typology as (i) it is not derived from existing work on embodied cognition more generally and serve the specific purpose of grounding (artificial) consciousness as one aspect of cognition, and because (ii) it would only warrant little differentiation of existing work; most \ac{CC} systems presently fall into a gap between the 1st and 2nd type.

Crucially, the extended typology is only loosely hierarchical. \emph{Historically} as well as \emph{physically} and \emph{virtually} embodied systems are all \emph{structurally coupled}. \emph{Physically} and \emph{virtually} embodied systems in turn can, but do not have to be \emph{historically} embodied. \emph{Organismoid} systems can be \emph{virtually} or \emph{physically} embodied, but we reserve \emph{organismic} embodiment to \emph{physically} embodied systems, as autonomy via organisational closure relies on physical forces that pose a real threat to a system's organisation and boundary.

\section{Review Method}

To investigate how embodiment has been discussed in \ac{CC} research, we performed a systematic literature review on the proceedings of the \acf{ICCC}, the prime venue for \ac{CC} research, between its inception in 2010 and its latest edition in 2020. We expect our findings to be representative, as \ac{ICCC} gathers a wide audience of CC researchers and practitioners, and welcomes contributions covering any creative domain, creative practice and aspect of creative cognition \citep{icccCfP}. We constrained our review to paper candidates that explicitly mention the words \enquote{embodiment}, \enquote{embodied}, \enquote{disembodiment}, \enquote{disembodied}, \enquote{embody}, or \enquote{embodying}. 

We acknowledge that our reliance on the explicit usage of the word may overlook a large amount of potentially relevant papers. This particularly concerns work on robotics, which often does not include explicit mentions of the embodiment of the investigated systems. Simply including the term \enquote{robotics} in our search however was impractical, as, for fairness, it would have required to also include any other type of system characterised by the remaining embodiment types. Given the inclusiveness of some types, the amount of potentially related papers would likely go beyond what could reasonably be reviewed in depth. Moreover, we believe that our present approach allows us to identify intentional and thus more informative discussions of the relationship between embodiment and (the perception of) creativity.

For our initial candidate paper selection, we divided the past \ac{ICCC} proceedings into {\it pdf} files, each containing one paper, short paper, demo description, or other peer reviewed publication. We used a wildcard search in Adobe Acrobat reader with the phrase \enquote{*embod*} and the hyphenated version \enquote{*em-bod*}. In total we found 99 papers mentioning the word \enquote{embodiment}, \enquote{embodied}, \enquote{disembodiment}, \enquote{disembodied}, \enquote{embody}, or \enquote{embodying} explicitly, with a total of 491 matches to the search phrases.

\begin{figure}[t]
  \includegraphics[width=1\columnwidth]{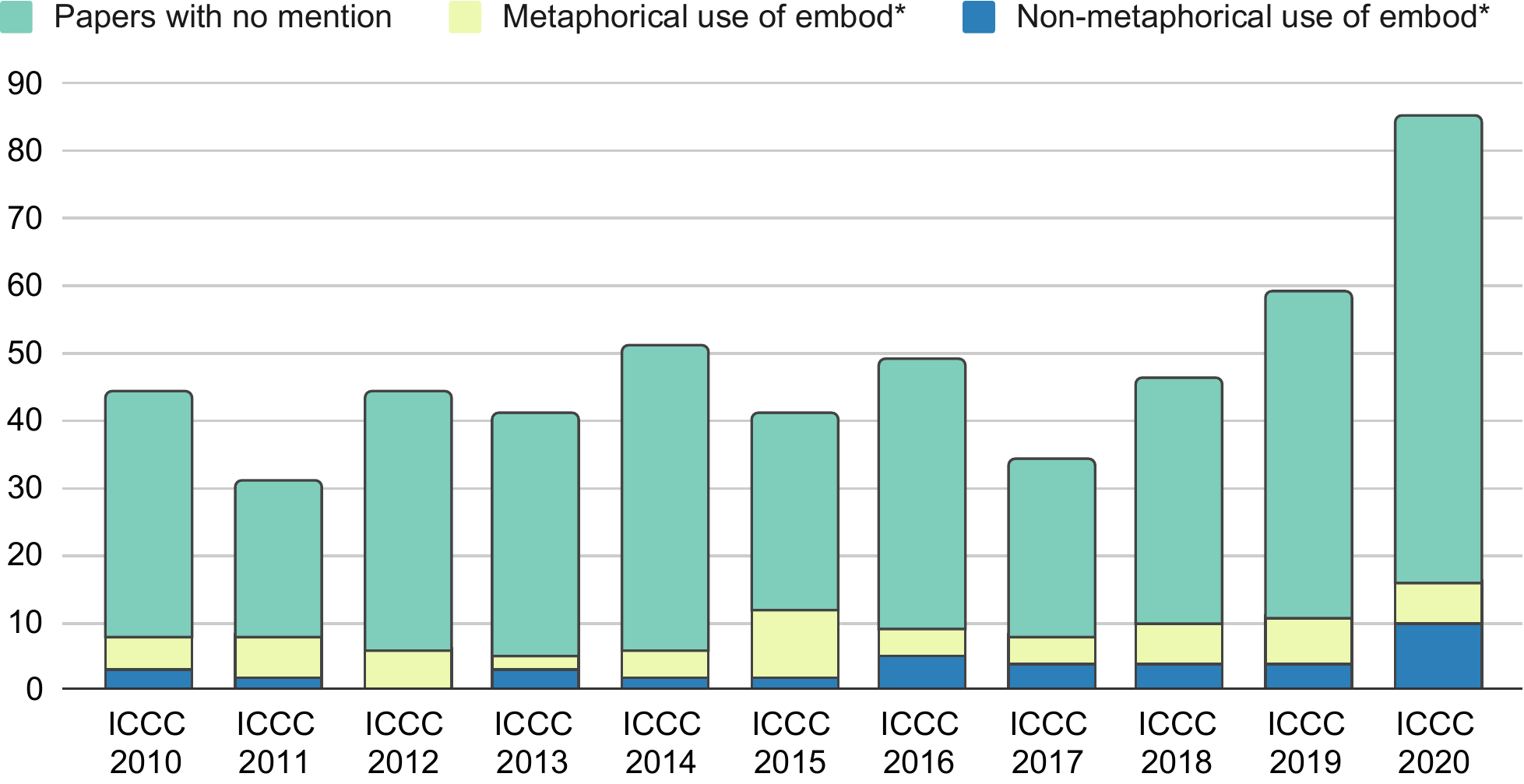}
  \caption{Absolute annual numbers of \ac{ICCC} papers using words derived from \enquote*{embodiment} in a (non-)metaphorical way. On average, 7.1\% of papers use the word \enquote*{embodiment} or its derivatives non-metaphorically, with the lowest proportion (0\%) in 2012 and the highest (11.8\%) in 2017 and 2020. Overall the term is used throughout the proceedings, but non-metaphorical usage has increased over the years.\vspace{-0.3cm}}
  \label{papers}
\end{figure}

We then reduced the candidate papers for our final, qualitative analysis in a two-stage process. We first excluded papers which mentioned these words only in the References section (7 papers), or as part of a general list of \ac{CC} related topics (2 papers). We secondly excluded papers that used these keywords in a merely metaphorical way, such as suggesting that a specific algorithm or system \enquote{embodies} certain values (51 papers). This left us with a final pool of 40 papers for in-depth analysis. Figure \ref{papers} illustrates the overall usage of the term over the years, thus partly answering \textbf{RQ1}.

We assessed the embodiment of any system described in the remaining papers, either introduced there or through reference to other work, based on Ziemke's \citeyearpar{ziemke2003s} extended typology. Since explicit definitions of embodiment were mostly absent, this usually required us to look at the specific characteristics of the system in question. To answer our research questions, we also gathered notes on:
\begin{itemize}
    \item Which challenges did we encounter in assessing the embodiment described in the paper?
    \item Did we identify any embodiment types that were not yet present in our typology? (\textbf{RQ1})
    \item Why have the authors embraced this particular type of embodiment, and what challenges did they face? (\textbf{RQ2})
    \item What did the paper express about the relationship between embodiment and creativity, or between embodiment and the perception of creativity? (\textbf{RQ3})
\end{itemize}

Each of the remaining 40 papers was first assessed by one of the four researchers participating in the effort. We then discussed the challenges encountered, and cross-checked each paper with another colleague in the team. Any potential disagreements were resolved in dedicated discussions. Another four papers were removed during this step because the explicit mentions to embodiment were cursory. This reduced the final number of papers to 36.

The answers to the questions presented above were analysed by building an \emph{affinity diagram}, a \emph{qualitative} data analysis method used to group data into emergent categories \cite[p.~286]{rogers11}.
We first built individual affinity diagrams from the answers collected for each question by grouping similar items together and labelling them. We then merged categories between the different questions, yielding one large affinity diagram as a connected view of different aspects of embodiment in \ac{CC}. We finally identified overarching themes within this structure.

\begin{table*}[t!]
        \centering
        \resizebox{0.93\textwidth}{!}{
		\begin{tabular}{llllllllllll}
			& & & & & \multicolumn{7}{c}{Embodiment Type}\\ \cmidrule(lr){6-12}\\[-0.3cm]
			Authors & Year & Abbreviated Title &  \rot{Concrete System(s)} & \rot{Mult. Embodiments} & \rot{Struct. Coupling} & \rot{Historical} & \rot{Virtual} & \rot{Physical} & \rot{Organismoid} & \rot{Humanoid} & \rot{Organismic}\\ \midrule
			
de Melo \& Gratch	&	2010	&	Evolving Expression of Emotions	 & \textemdash & \textemdash & \LEFTcircle & \Circle & \LEFTcircle & \Circle & \LEFTcircle & \LEFTcircle & \Circle \\

Saunders et al.	&	2010	&	Curious Whispers	 & \checked & \textemdash & \CIRCLE & \CIRCLE & \Circle & \CIRCLE & \LEFTcircle & \Circle & \Circle\\

&	& &  & &  &  &  &  &  &  \\[-1.5ex]

Kirsh	&	2011	&	Creative Cognition in Choreography	 & \textemdash & \textemdash &  \CIRCLE & \CIRCLE & \Circle & \CIRCLE & \CIRCLE & \CIRCLE & \CIRCLE\\

&	& &  & &  &  &  &  &  &  \\[-1.5ex]

Gemeinboeck  \& Saunders	&	2013	&	Creative Machine Performance: Robotic Art	 & \checked & \textemdash & \CIRCLE & \CIRCLE & \Circle & \CIRCLE & \LEFTcircle & \Circle & \Circle\\

Saunders, Chee  \&  Gemeinboeck	&	2013	&	Evaluating Human-Robot Interaction	 & \checked & \textemdash & \CIRCLE & \CIRCLE & \Circle & \CIRCLE & \LEFTcircle & \Circle & \Circle\\

Schubert  \& Mombaur	&	2013	&	Motion Dynamics in Abstract Painting	 & \checked & \textemdash & \CIRCLE & \LEFTcircle & \Circle & \CIRCLE & \Circle & \Circle & \Circle\\

&	& &  & &  &  &  &  &  &  \\[-1.5ex]

Schorlemmer et al.	&	2014	&	COINVENT: Concept Invention Theory	 & \textemdash & \textemdash & \Circle & \Circle & \Circle & \Circle & \Circle & \Circle & \Circle \\

Davis et al.	&	2014	&	Building Artistic Computer Colleagues	 & \checked & \textemdash & \CIRCLE & \Circle & \Circle & \LEFTcircle & \Circle & \Circle & \Circle\\

&	& &  & &  &  &  &  &  &  \\[-1.5ex]

Jacob  \& Magerko	&	2015	&	Interaction-based Authoring	 & \checked & \textemdash & \CIRCLE & \CIRCLE & \CIRCLE & \Circle & \CIRCLE & \LEFTcircle & \Circle \\

Takala	&	2015	&	Preconceptual Creativity	 & \checked & \textemdash & \CIRCLE & \CIRCLE & \CIRCLE & \Circle & \Circle & \Circle & \Circle \\

&	& &  & &  &  &  &  &  &  \\[-1.5ex]

Carlson et al.	&	2016	&	Cochoreo: Generative Choreography	 & \checked & \textemdash & \CIRCLE & \Circle & \CIRCLE & \Circle & \Circle & \Circle & \Circle\\

McCormack  \&  d’Inverno	&	2016	&	Designing Improvisational Interfaces	 & \checked & \textemdash  & \CIRCLE & \Circle & \Circle & \Circle & \Circle & \Circle & \Circle\\

Crnkovic-Friis  \& Crnkovic-Friis	&	2016	&	Generative Choreography using Deep Learning & \checked & \textemdash  & \Circle & \Circle & \Circle & \Circle & \Circle & \Circle & \Circle \\
Guckelsberger et al. &	2016 &	Supportive and Antagonistic Behaviour	 & \checked & \checked & \CIRCLE & \CIRCLE & \CIRCLE & \CIRCLE & \LEFTcircle & \Circle & \Circle \\

Brown	&	2016	&	Understanding Musical Practices	 & \checked & \textemdash & \CIRCLE & \Circle & \Circle & \Circle & \Circle & \Circle & \Circle \\

&	& &  & &  &  &  &  &  &  \\[-1.5ex]

Augello et al.	&	2017	&	Creative Robot Dance With Variational Encoder	 & \checked & \textemdash & \CIRCLE & \CIRCLE & \CIRCLE & \Circle & \CIRCLE & \CIRCLE & \Circle \\

Fitzgerald, Goel \& Thomaz	&	2017	&	Human-Robot Co-Creativity: Task Transfer & \checked & \textemdash & \CIRCLE & \CIRCLE & \Circle & \CIRCLE & \CIRCLE & \CIRCLE & \Circle \\

Guckelsberger, Salge \& Colton	&	2017	&	Non-Anthropocentric Intentional Creative Agency	 & \checked & \checked & \CIRCLE & \CIRCLE & \Circle & \CIRCLE & \CIRCLE & \CIRCLE & \CIRCLE \\

Singh et al. &	2017	&	Unified Classification and Generation Networks	 & \checked & \textemdash & \CIRCLE & \Circle & \CIRCLE & \Circle & \Circle & \Circle &\Circle \\

&	& &  & &  &  &  &  &  &  \\[-1.5ex]

Colton, Pease \& Saunders	&	2018	&	Issues of Authenticity in Creative Systems	 & \checked & \checked & \CIRCLE & \CIRCLE & \Circle & \CIRCLE & \LEFTcircle & \Circle & \Circle \\

Hedblom et al. &	2018	&	Under the Super-Suit: Conceptual Blending	 & \checked & \textemdash & \Circle & \Circle & \Circle & \Circle & \Circle & \Circle & \Circle\\

Saunders \& Gemeinboeck	&	2018	&	Performative Body Mapping for Robots	 & \checked & \checked & \LEFTcircle & \Circle & \Circle & \LEFTcircle & \LEFTcircle & \Circle & \Circle \\

Wicke \& Veale	&	2018	&	Interview With the Robot	 & \checked & \textemdash & \CIRCLE & \LEFTcircle & \Circle & \CIRCLE & \CIRCLE & \CIRCLE & \Circle \\

&	& &  & &  &  &  &  &  &  \\[-1.5ex]

Jacob et al.	&	2019	&	Affordance-Based Generation of Object Variants	 & \checked & \textemdash & \Circle & \Circle & \CIRCLE & \Circle & \Circle & \Circle & \Circle\\

Onate, Mendez \& Gervas	&	2019	&	Elevator Conversations Based on Emotions	 & \checked & \textemdash & \Circle & \Circle & \Circle & \Circle & \Circle & \Circle & \Circle\\

Veale, Wicke \& Mildner	&	2019	&	Duets Ex Machina: “Double Acts”	 & \checked & \checked & \CIRCLE & \Circle & \Circle & \CIRCLE & \CIRCLE & \CIRCLE & \Circle \\

&	& &  & &  &  &  &  &  &  \\[-1.5ex]

Loesel, Mirowski \& Mathewson	&	2020	&	Do Digital Agents Do Dada?	 & \checked & \checked & \LEFTcircle & \Circle & \CIRCLE & \LEFTcircle & \LEFTcircle & \LEFTcircle & \LEFTcircle \\

Alexandre	&	2020	&	Creativity Explained by Cognitive Neuroscience	 & \checked & \textemdash & \CIRCLE & \CIRCLE & \Circle & \CIRCLE & \CIRCLE & \CIRCLE & \CIRCLE\\

Chen et al.	&	2020	&	Breaking the Imitation Barrier & \checked & \textemdash & \Circle & \Circle & \Circle & \Circle & \Circle & \Circle & \Circle\\

Ferguson et al.	&	2020	&	Automatic Similarity Detection in LEGO Ducks	 & \checked & \textemdash & \Circle & \Circle & \Circle & \Circle & \Circle & \Circle & \Circle\\

Kantosalo et al.	&	2020	&	Modalities, Styles and Strategies	 & \textemdash & \textemdash & \CIRCLE & \Circle & \Circle & \CIRCLE & \Circle & \Circle & \Circle\\

Kantosalo \& Takala	&	2020	&	Five C’s for Human–Computer Co-Creativity	 & \checked & \checked & \CIRCLE & \Circle & \Circle & \CIRCLE & \Circle & \Circle & \Circle\\

LaViers \& Vidrin	&	2020	&	Can a Robot Do a Trust Fall? & \textemdash & \textemdash & \CIRCLE & \Circle & \Circle & \CIRCLE & \Circle & \Circle & \Circle\\

Savery, Zahray \& Weinberg	&	2020	&	Shimon the Rapper: Robot Rap Battles	 & \checked & \textemdash & \CIRCLE & \LEFTcircle & \Circle & \CIRCLE & \CIRCLE & \CIRCLE & \Circle \\

Wallace et al.	&	2020	&	Movement Generation with Audio Features	 & \checked & \textemdash & \Circle & \Circle & \Circle & \Circle & \Circle & \Circle & \Circle\\

Wicke \& Veale	&	2020	&	Show, Don’t (Just) Tell	 & \checked & \textemdash & \LEFTcircle & \Circle & \Circle & \CIRCLE & \CIRCLE & \CIRCLE & \Circle \\

		\end{tabular}
	}
	\caption{Chronological overview of \ac{ICCC} (2010-2020) papers mentioning the concept of embodiment explicitly and non-metaphorically. The circles \Circle\, \LEFTcircle, and \CIRCLE\ represent increasing degrees to which the described embodiment(s) match our types. Organismoid embodiment entails humanoid embodiment, but we also discriminate humanoid embodiment separately.	For papers describing multiple embodied systems, e.g.~a robot and a human, individual embodiment types were combined with a logic \enquote{or}. Papers introducing a “concrete system” have been marked; other papers consider theory or abstract systems.\vspace{-0.3cm}}
\label{tbl:embodiment_papers}
\end{table*}

\section{Findings}
We report our findings on the state-of-the-art of embodied \ac{CC} in individual subsections. We first detail the types of embodiment identified in existing work, thus conclusively answering \textbf{RQ1}. We elaborate on our difficulties in this process later in the discussion section, where we go beyond the present findings and make recommendations for future research. Our review uncovered that existing insights on the relationship of embodiment and (the perception of) creativity are often tightly aligned with researchers' motivations to embrace a certain type of embodiment in their work. In the second part of our findings, we consequently address \textbf{RQ2} and \textbf{RQ3} jointly through themes corresponding to \emph{opportunities} and \emph{challenges} for embodied \ac{CC}. We distinguish each theme as a paragraph heading, point out sub-themes in \textbf{bold}, and highlight which embodiment type it relies on in \emph{italics}.

\subsection{Embodiment Types}
Table \ref{tbl:embodiment_papers} provides an overview of our final paper selection, together with our assessment of the described embodiment. 
We found that \emph{structural coupling} and \emph{physical embodiment} are most common, each appearing in almost twenty papers. \emph{Historical}, \emph{virtual} and \emph{organismoid} mentions appear quite equally, each in about ten papers. \emph{Organismic embodiment} was only identified in four papers; \citet{Guckelsberger2017a} discuss it with respect to machines, while the other three instances relate to humans, which are organismically embodied by definition. Over the years, although there is considerable variation, the annual number of papers grows from 1.6 on average in the first half to 3.6 in the latter half of the decade, until there is a sudden peak of 10 in 2020. Mentions of different types follow a similar trend, except that \emph{structural coupling} gives way to the more specific, \emph{physical embodiment} in recent years.
Out of the 36 papers, 23 present a concrete, embodied computational system, seven are at least partly theoretical or appealing to the embodiment of humans. In another seven papers, we were unable to identify one or multiple types of embodiment at all, indicated by a row of empty circles in Table \ref{tbl:embodiment_papers}.

Our analysis highlights \emph{virtual} and \emph{physical embodiment} as strongest differentiators of existing work. It moreover identified a chasm between researchers poised to leverage \emph{humanoid embodiment}, and those rejecting it for the benefits of other variations of \emph{organismoid embodiment}.

Through our review, we identified one embodiment type that was not directly present in our extension of Ziemke's \citeyearpar{ziemke2003s} typology: \citet{loesel2020} introduce \enquote{cyborg embodiment}, which bridges between \emph{virtual}, \emph{organismoid (anthropomorphic)} and \emph{physical embodiment}. It was demonstrated in their theatrical experiment \enquote{AI Improv}, where a chatbot provides sentences to be articulated by a human actor, and receives new prompts to react to from a backstage operator who monitors the on-stage dialogue. The two people thus provide a split actuation and sensing interface to the artificial system, allowing for it to be embedded in the physical environment of the stage. 

\subsection{Opportunities of Embodied \ac{CC}}

Our analysis identified nine opportunities of embracing embodiment in CC. We can distinguish two sub-groups, based on how the themes relate to the concept of interaction. All types of embodiment distinguished in our typology assume at least a minimal form of interaction between an agent and its environment in the form of \emph{structural coupling}. The first four themes concern how this embodiment-induced split provides opportunities for modelling a specific creative domain, outsourcing computation, letting creativity emerge and stimulating it. The remaining five themes operate on a stronger notion of interaction between agents, i.e.~where the embodied agent's environment comprises interaction partners. In this group, embodiment is considered a means to model co-creativity, ground meaning, facilitate more natural interaction with people, support identification and empathy with the computational agent, and increase the \ac{CC} system's creative intentionality and autonomy.

\vspace{-0.3cm}\paragraph{Domain Necessity}
\emph{Any type of embodiment} presents the opportunity \textbf{to model creative processes that unfold between an agent and their environment}. Some creative domains may necessitate this split more than others in order to comprehensively model the creative processes within. Dance choreography represents a prime example \citep{augello2017, carlson2016cochoreo}, but embodiment has also been embraced in e.g.~music \citep{SchorlemmerSKKC14} and painting \citep{schubert2013role, singh2017unified} to model creative processes that rely on sensorimotor feedback between an agent and their surroundings.

\vspace{-0.3cm}\paragraph{Outsourcing Computation}
Embodied \ac{CC} has adopted several premises of embodied AI more generally, notably the use of \emph{physical embodiment} to \textbf{outsource computation into the physical world}. \citet{saunders2013evaluating} note that physical embodiment allows artificial agents \enquote{to take advantage of properties of the physical environment that would be difficult or impossible to simulate computationally} (paraphrasing \citeauthor{brooks1990elephants}, \citeyear{brooks1990elephants}; see also \citeauthor{gemeinboeck2013creative}, \citeyear{gemeinboeck2013creative}). They thus relate to one of the most prominently articulated benefits of embodied AI: the use of the world \enquote{as its own model}~\citep{brooks1991intelligence}.

\vspace{-0.3cm}\paragraph{Emergent Creativity}
While the outsourcing of computation seems a mere engineering benefit at first, it has major implications for the creativity that a \emph{physically embodied} system can potentially exhibit. In particular, it allows to realise the very premise of systems theories of creativity \citep{csikszentmihalyi1988society}, the \textbf{emergence of creative behaviour through an agent's interaction with their environment}, including other agents: \enquote{embodiment provides opportunities for agents to experience the emergence of effects beyond the computational limits that they must work within} \citep{saunders2013evaluating}. In the art installation \emph{Zwischenräume}, the robots' creative agency \enquote{is not predetermined but evolves based on what happens in the environment they examine and manipulate} \citep{gemeinboeck2013creative}. This emergence benefits from a controller that is not pre-coded but sensitive to an agent's changing embodiment. Several authors \citep[e.g.][]{saunders2010curious, saunders2013evaluating, Guckelsberger2016a, Guckelsberger2017a} highlight the use of computational intrinsic motivation to this end. A system with a suitable controller can leverage its embodiment to \textbf{expand its behavioural range beyond what can be anticipated by the system designer}, realising novelty as core criterion for creativity \citep{rhodes1961}.

\vspace{-0.3cm}\paragraph{Stimulation of Creativity}
As a specific case of emergent creative behaviour, several researchers argue that \textbf{constraints imposed through embodiment can stimulate creativity}. Drawing on \citet{pickering2005embodiment}, Saunders et al.~highlight that the \enquote{world offers opportunities, as well as presenting constraints: human creativity has evolved to exploit the former and overcome the latter, and in doing both, the structure of creative processes emerge} \citep{saunders2010curious}. \citet{Guckelsberger2016a} argue in an artistic context, drawing on different embodied \ac{CC} systems and a thought-experiment on the \emph{physically embodied} robot society \emph{Curious Whispers} \citep{saunders2010curious}, that overcoming embodiment-related constraints in an environment can necessitate and -- given a suitable agent controller -- yield creativity. \citet{takala2015preconceptual} demonstrates this on a simulated robotic arm capable of inventing new and useful movements when encountering obstacles. This also highlights that creative action is possible without creative reasoning, a distinction which is later picked up by \citet{fitzgerald2017human}. By utilising \emph{virtual embodiment}, Takala demonstrates that the effect of embodiment constraints on creativity can be investigated without physical embodiment. However, the use of physical embodiment can better alleviate doubts about the emergent behaviour being truly novel, and not engineered a priori into a simulated environment.

\vspace{-0.3cm}\paragraph{Co-Creativity}
Many of the analysed papers express a focus on stronger forms of interaction between agents. Two particular modes of interaction are given by human-machine and machine-machine co-creativity~\citep{saunders2015computational,kantosalo2016modes}. This focus can be explained with the observation that \textbf{embodiment is a prerequisite for co-creativity}. \citet{Guckelsberger2016a} highlight that \enquote{co-creative and social creativity systems are only meaningful if each agent has a different perspective on a shared world, allowing them to complement each other, and for creativity to emerge from their interaction}. The necessary separation of agent and environment is crucially facilitated by \emph{any type of embodiment}. This allows for the attribution of embodiment to a given system, based on the systemic nature of the system alone. The next three themes represent additional lenses on human-machine co-creativity.

\vspace{-0.3cm}\paragraph{Grounding Meaning}
Another central premise of more radical theories of embodied cognition, bordering to enactivism~\citep{varela2017embodied}, is that \emph{physical embodiment} can \textbf{overcome symbolic representations and ground meaning in sensorimotor interaction} \citep[e.g.][]{dreyfus1992computers}. \citet{colton2018issues} stress that this allows \textbf{re-representing} creative domains in action. As an example, they refer to the Marimba playing robot \emph{Shimon} \citep{hoffman2010shimon} which represents music as choreography of physical gestures. They also emphasise the grounding of machine \enquote{life experiences} as an important factor in \textbf{increasing the perception of authenticity in \ac{CC} systems}. Related, Wicke, Veale and Mildner exploit robot gestures to provide the illusion of grounding computer-generated stories (Wicke and Veale, \citeyear{wicke2018interview}, \citeyear{wicke2020}; \citeauthor{veale2019duets}, \citeyear{veale2019duets}), thus leveraging embodiment to affect the perception of their robot's creativity.

\vspace{-0.3cm}\paragraph{Natural Interaction}
\citet{saunders2010curious} are first to stress that \emph{physical embodiment} allows for \ac{CC} systems to be \textbf{embedded in rich social and cultural environments}. This enables \enquote{computational agents to be creative in environments that humans can intuitively understand} \citep{saunders2013evaluating}. Robotic art installations are highlighted as one means to \enquote{gain access to shared social spaces with other creative agents, e.g., audience members} \citep{gemeinboeck2013creative}. Existing research often stresses that situating \ac{CC} systems in physical space \textbf{realises more natural interaction} by established means, and can \textbf{unleash new modes of interaction}. This is explained by \emph{physical embodiment} \textbf{affording tight feedback loops} \citep{wicke2018interview} and \textbf{providing stronger cues} to the human interaction partners \citep{saunders2013evaluating}. In some instances, this interaction is constrained to a few invididuals, e.g.~when situating a robot on stage to interact with musicians \citep{savery2020} or actors \citep{loesel2020}. In their art installation \emph{Zwischenräume}, \citeauthor{gemeinboeck2013creative} in contrast open the interaction to a wider audience in an exhibition space, permitting \enquote{the development of significantly new modes of interaction} and \enquote{engaging a broad audience in the questions raised by models of artificial creative systems} \citep{gemeinboeck2013creative}. Within natural interaction spaces, \emph{physical embodiment} can \enquote{improve the relationship between humans and AI, inspiring humans in new creative ways} \citep{savery2020}, e.g.~in partnering a human musician with \emph{Shimon's} reincarnation as rapper.

\vspace{-0.3cm}\paragraph{Identification \& Empathy}
Within the overarching theme of agent interaction, researchers embraced \emph{physical}, and in particular \emph{organismoid} and \emph{humanoid embodiment} to \textbf{facilitate and improve communication} and, consequently, to \textbf{afford identification, empathy and affect} between human and robot, or within a society of robots. \citeauthor{gemeinboeck2013creative} highlight embodied action in their installation \emph{Zwischenräume} as a means for communication; it takes the form of robots creating noises with a hammer which members of an exhibition audience and other robots can perceive and react to. They moreover stress from an enactivist perspective that the robots' actions provide \enquote{a window on the agents’ viewpoint} \citep{gemeinboeck2013creative}, thus possibly facilitating more introspection. \citeauthor{wicke2018interview} refer to work outside \ac{CC} to emphasise that, \enquote{when identification with the [story]teller is the goal, the physical presence of a moving body with a human shape makes all the difference} \citep{wicke2018interview}, hence stressing the effect of \emph{organismoid (humanoid) embodiment}. Moreover, they hypothesise that a \enquote{listener that can identify with the storyteller is better positioned to empathize with the story that the teller wants to convey, especially when that story is crafted from the life experiences of the listeners themselves}. This reliance on life experience resonates with Colton, Pease and Saunder's \citeyearpar{colton2018issues} previously mentioned factors to improve the authenticity of \ac{CC} systems.

\vspace{-0.3cm}\paragraph{Intentionality \& Autonomy}
While \emph{organismic embodiment} is rarely addressed in the literature and typically only through human embodiment \citep{kirsh2011creative,alexandre2020creativity, loesel2020}, \citet{Guckelsberger2016a} highlight that \emph{organismic embodiment} realised in artificial systems might play a central role in future \ac{CC} research. Their theoretical investigation sets out by considering non-artistic creativity in simple computational systems through the lens of autopoietic enactivism \citep{maturana1987tree} as adopted in the theory of enactive AI \citep{froese2009enactive}. The latter theory holds that machines with organismic embodiment can, similar to living beings, realise an intrinsic purpose by maintaining the precarious existence induced by this form of embodiment. It moreover claims that this intrinsic purpose can ground intentional agency. \citet{Guckelsberger2016a} consider more specifically when organismic embodiment can ground \textbf{intentional creative agency}, realising genuine creative autonomy \citep{jennings2010developing}. They argue that a machine grounds value and novelty in creative activity through the maintenance of their precarious identity, based on acts of self-production and adaptation against entropic forces. The claim that intentional creative agency is contingent on \emph{organismic embodiment} allows for additional, radical statements. \citet{Guckelsberger2016a} argue via \citet{dreyfus2007heideggerian} that a \ac{CC} system might have to accurately reproduce human \emph{organismic embodiment} to \textbf{exhibit human-like creativity with intentional agency}. Moreover, they introduce the concept of \enquote{embodiment distance} to put forward hypotheses on the impact of \emph{organismic embodiment} on the perception of creativity: \enquote{when we evaluate the creativity of non-human systems with intentional agency, we are likely to misjudge value in their behaviour or artefacts, or hesitate to attribute any value at all, as our embodiment distance is too large}~\citep{Guckelsberger2016a}. This makes the mimicking of \emph{organismoid}, and potentially, \emph{humanoid embodiment} relevant. \citet{colton2018issues} extend the concept of embodiment distance to non-organismically embodied systems and discuss ways to overcome it to foster the perception of creativity and authenticity in \ac{CC} systems.

\subsection{Challenges of Embodied \ac{CC}}
Our analysis exposed that many opportunities for embracing embodiment in \ac{CC} have a flip-side, the impact of which is mediated by the respective embodiment type. The identified three challenges are easily overlooked, as they are often addressed separately from the corresponding opportunities.

\vspace{-0.3cm}\paragraph{Computational \& Design Costs}
While affording a range of opportunities, such as more natural interaction, the stimulation of creativity, grounding, etc. -- being embedded in our physical world also puts \textbf{high demands on [the] hardware, software and system engineering} \citep{saunders2010curious} of \emph{physically embodied} agents. \citet{fitzgerald2017human} particularly lament the increased processing costs due to the high dimensionality of robot sensors and actuators. \citeauthor{gemeinboeck2013creative} summarise that \enquote{embodying creative agents and embedding them in our everyday or public environment is often messier and more ambiguous than purely computational simulation} \citep{gemeinboeck2013creative}. Especially when tempted by the opportunity to outsource computation into our physical environment, these costs must be carefully weighted.

\vspace{-0.3cm}\paragraph{Unpredictability}
Related, creative behaviour that emerges from the interaction of \emph{any embodied} agent and their environment, especially if resulting from intrinsic motivation, is often \textbf{hard to predict} \citep{Guckelsberger2016a}. This is more relevant in some application domains than others, with many artistic domains affording unique possibilities for playful experimentation. Across domains however, researchers must exercise particular caution when designing for interaction with people. Crucially, \emph{virtual embodiment} comes with more well-defined interaction interfaces and affords more control in experiments that can be reset and afford stronger introspection. 

\vspace{-0.3cm}\paragraph{False Expectations}
Several authors express hope that \emph{organismoid}, in particular \emph{humanoid embodiment} can facilitate stronger identification, empathy and affect \citep[e.g.][]{wicke2018interview}. \citeauthor{saunders2018performative} however warn that \textbf{humanoid robots can cause disappointment} as they \enquote{elicit human investment based on superficial and often false social cues} \citep{saunders2018performative}. Referencing studies of human-robot interaction \citep{dautenhahn2013}, they particularly highlight the risk of shaping human expectations in a robot's social capabilities based on appearance alone. As a workaround, they suggest focusing research efforts on non-anthropomorphic robots, and on generating embodied natural movement as means of identification, instead of a similar form or sensorimotor equipment.

\section{Discussion of Embodiment Assessment}

We faced several challenges in assessing the type of embodiment in the selected papers. We briefly elaborate on embodiment (i) classification challenges, (ii) biases, (iii) \enquote{under-} and (iv) \enquote{over-attributions}, and (v) typology limitations.

The classification (i) of embodiment was complicated by unspecific descriptions especially in theoretical papers. Moreover, some papers related to several systems at once, or exclusively addressed human rather than machine embodiment. Sometimes the lack of specifics did not allow us to gain insights into the use of embodiment in \ac{CC}. \citet{SchorlemmerSKKC14} for instance only appeal to embodied cognition in a side-note. Affected contributions are listed in Table  \ref{tbl:embodiment_papers} without an assessment of their embodiment type.

Biased views of embodiment (ii) were expressed when authors only explicitly recognised embodiment in humans, e.g.~in the case of a virtual system affecting the embodiment of a human user, or when discussing the challenges of modelling human movement. An example of this can be seen in \cite{schubert2013role}, who attempt to capture the movement dynamics of human painters.

Some authors \enquote{under-attributed} (iii) embodiment in that they leveraged a specific embodiment without explicitly referring to its type or properties. We observed this particularly often for \emph{organismoid embodiment}, e.g.~in Saunders et al.'s \citeyearpar{saunders2010curious, saunders2013evaluating} description of the \emph{Curious Whispers} robot society. Here, the robots' bug-like, organismoid embodiment is not discussed explicitly, although it may have a specific effect on the perception by a human interaction partner.

Closely related, other authors \enquote{over-attributed} (iv) embodiment in that they explicitly appealed to a certain type of embodiment without fully implementing or discussing its required properties. For example \citeauthor{DBLP:conf/icccrea/MeloG10} (\citeyear{DBLP:conf/icccrea/MeloG10}) describe an experiment performed on \enquote{virtual humans}, but they are used as passive mannequins to shine light on. Given that a simulated environment to perturb and be perturbed is absent, we cannot even attest \emph{structural coupling}. Again, this was particularly evident for \emph{organismoid embodiment}.

Limitations of our typology (v) made it challenging to identify specific embodiment types, in particular \emph{historical embodiment}: several systems presented memory or learning capabilities, but this learning did not necessarily happen during the systems' lifetime. This launched a discussion amongst the authors on whether e.g.~the existence of learning hardware is indicative of historical embodiment. We eventually agreed that historical embodiment is independent of a lifetime criterion or learning, and present if the system's past structural coupling has an effect on its future coupling, e.g.~when a past perception triggers a change in a virtual or physical sensor, e.g. affecting a camera's angle, thus influencing future perceptions. \emph{Organismoid embodiment} was also difficult to assess, as the proximity of a system in shape and sensorimotor equipment to living beings is a continuum. Given the focus of many systems on humanoid embodiment, it would have been helpful to consider it a separate type.

Some types of embodiment turned out to be worse at differentiating and distinguishing existing work and insights than others, but this does not necessarily disqualify them from future use. \emph{Structural coupling} for instance is the most inclusive type of embodiment and applied to almost all reviewed systems, in particular to any physical object; yet, it proved valuable for sanity-checking whether a simulated system can be considered embodied at all. \emph{Organismic embodiment} in contrast was the most exclusive type. We think it should be included nonetheless, given that existing theoretical work \citep{Guckelsberger2017a} assigns it a major role in future \ac{CC} research, e.g.~on creative autonomy. As for any other type, the current lack of differentiating power might indicate an under-appreciation in existing work, rather than a weakness of the type itself.

\section{Directives for Embodied \ac{CC} Research}

Our analysis supports the hypothesis which motivated this systematic review in the first place: creativity in artificial systems, and how it is perceived, is affected by embodiment. If we assume this hypothesis, then furthering our insights into embodied \ac{CC} will play a critical role in advancing the goals of \ac{CC} more generally. However, existing research is highly fragmented and ambiguous, lacks generalising empirical results, and rarely trades-off the opportunities and challenges of a certain type of embodiment. We translate findings from our review on \textbf{RQ1}, \textbf{RQ3} and \textbf{RQ2} into three directives as pillars of a future embodied \ac{CC} research agenda. 

\vspace{-0.3cm}\paragraph{Clarify embodiment:} Our review highlighted that many opportunities and challenges of embodied \ac{CC} are exclusive to a specific type of embodiment. However, we experienced serious difficulties in assessing the specific type of embodiment embraced in existing work, as documented in the corresponding papers. In order to establish an efficient, unambiguous, verbal and written scientific discourse on embodied \ac{CC}, we urge researchers to always clarify what specific embodiment they appeal to in a particular theoretical or applied project. To this end, adopting definitions from typologies as presented here may serve as a shortcut, foster comparison, and alleviate the \enquote{under-} or \enquote{over-attribution} of a certain embodiment. However, any typology can be contested and lacks detail; references to specific embodiment types should thus always be complemented with extensive descriptions.

To counteract the fragmentation of research output, we recommend to reference embodiment-related work extensively. This review, albeit non-exhaustive, might serve as a starting point to identify relevant work.

\vspace{-0.3cm}\paragraph{Conduct empirical studies:} Our review moreover uncovered that no existing \ac{CC} study produced generalising, empirical insights about the effect of embodiment on (i) an artificial system's creativity, and (ii) how its creativity is perceived by others, including humans. Researchers either make assumptions on this relationship, or draw on existing empirical findings from other fields that may not easily translate to computational systems \citep[e.g.][]{brown2016understanding}. We recommend to conduct qualitative and quantitative empirical studies on the impact of a specific embodiment, treated as independent variable, on (the perception of) creativity. Informed by our analysis of which embodiments differentiated existing work most, we recommend that initial empirical studies should investigate \emph{virtual} and \emph{physical}, or \emph{humanoid} and other variations of \emph{organismoid embodiment}, as values of the independent variable. 

In evaluating the perception of creativity (ii) as dependent variable, experimenters must eliminate or weight for creative ability (i). Vice versa, in evaluating creative ability (i), they must avoid bias by the perception of creativity (ii), e.g.~by employing objective measures of creativity \citep{ritchie2007some}. When employing subjective measures, researchers should consider concepts introduced in the literature such as the \emph{embodiment gap} \citep{Guckelsberger2017a, colton2018issues} between the system and its evaluator(s) as mediating variable.

\vspace{-0.3cm}\paragraph{Trade-off opportunities and challenges:} We encourage researchers to eventually make use of these empirical insights for the design of systems that reliably leverage a certain embodiment as a means to an end, e.g.~to accurately model creative processes that emerge from sensorimotor interaction in a specific domain. Crucially though, our review showed that most embodiment-related opportunities come with a challenging flip-side. In order to avoid an unfavourable trade-off, we recommend researchers to always inform their choice of a particular embodiment not only by its opportunities, but also by the corresponding challenges.

\section{Conclusions and Future Work}

Motivated by the potential impact of embodiment on creativity and its perception in artificial systems, we set out to map the present landscape of embodied \acf{CC}, and offered directions for future research in this area. To this end, we conducted a systematic review of papers presented at the \acf{ICCC} that explicitly discuss embodiment. To counteract ambiguity in the concept's use, we adopted a well-established embodiment typology, and extended it based on the recent scientific debate. 
We found that most existing work can be differentiated by its focus on \emph{virtual} vs.~\emph{physical}, and, more fine-granularly, on \emph{humanoid} vs.~\emph{non-humanoid, organismoid embodiment}. Moreover, we showed that each type comes with its unique opportunities and challenges as flip-sides of the same coin. Overall, we identified nine opportunities, e.g.~the outsourcing of computation or support for more natural interaction, and three challenges, e.g.~unpredictability and the shaping of false expectations, from existing studies. We identified several shortcomings of existing work that likely hinder progress on embodied \ac{CC} research, most prominently ambiguity in the use of the embodiment concept and a lack of dedicated empirical research. We leverage these insights in our final contribution: three directives to advance embodied \ac{CC} research.

While we chose the scope of this study to provide a reasonably unbiased, big-picture view of embodied \ac{CC}, future work should be dedicated to incorporating additional, relevant references. How to meaningfully constrain the scope is an open question, as relevant work can not only be found in be found in \ac{CC} books, but also related fields such as videogame AI, robotics, design and art. To allow for fairer and more direct, unambiguous comparisons, we moreover suggest to complement the present review methods with interviews, in which researchers in embodied \ac{CC} are asked directly to describe the embodiment in their concrete system or theory. We also deem it worthwhile to conduct a separate, deeper investigation of embodiment in co-creative systems and creative system societies, drawing on theories of embodied, embedded, extended and enactive social cognition \citep[e.g.][]{barsalou2003social, dautenhahn1997could,de2008making}, and on embodied interaction \citep[e.g.][]{dourish2001action} as well as embodied aesthetics research \citep[e.g.][]{scarinzi2015aesthetics}. Together with these extensions, our review and the extended typology could eventually benefit a longitudinal analysis of embodied \ac{CC}, identifying trends in the attitudes towards using specific embodiment types and their associated opportunities and challenges over time and across venues. We hope that this paper provides the necessary knowledge, inspiration, and guidance to drive future research on embodied \acf{CC}.

\section{Acknowledgements}
\small
We thank our reviewers for their valuable feedback, and Rob Saunders for commenting on the camera-ready paper. We are grateful to Marcus Scheunemann for advising us on social embodiment. CG is funded by the Academy of Finland (AoF) flagship programme \enquote{Finnish Center for Artificial Intelligence} (FCAI). AK \& TT are funded by the AoF project \enquote{Digital Aura} (\#311090). SNY is funded by CONACyT, project \enquote{ReNACE} (CB A1-S-21700).

\bibliographystyle{iccc}
\small
\bibliography{iccc}

\end{document}